\documentclass[runningheads]{llncs}

\usepackage[T1]{fontenc}
\usepackage{tabularx}
\usepackage{graphicx}
\usepackage{authblk}
\usepackage{xurl}

\newcommand{\equalcontrib}{\textsuperscript{+}}
\usepackage[ 
    colorlinks=false,
    pdfborder={0 0 0},
    filecolor=black,      
    urlcolor=black,
    linkcolor=black
]{hyperref}
\usepackage{float}
\usepackage{amssymb}
\usepackage{subcaption}
\usepackage[table]{xcolor}
\usepackage{collcell}
\usepackage{hhline}
\usepackage{amsmath}
\usepackage{pgf}
\usepackage{multirow}
\usepackage{array}
\def\colorModel{hsb} 

\newcommand\ColCell[1]{
  \pgfmathparse{#1<0?1:0}
    \ifnum\pgfmathresult=0\relax\color{white}\fi
  \pgfmathsetmacro\compA{0}     
\pgfmathsetmacro\compB{0} 
\pgfmathsetmacro\compC{#1/100}    
  \edef\x{\noexpand\centering\noexpand\cellcolor[\colorModel]{\compA,\compB,\compC}}\x #1
  } 
\newcolumntype{E}{>{\collectcell\ColCell}m{0.4cm}<{\endcollectcell}}

\usepackage{color}

\begin{document}

\title{Robust Alignment of the Human Embryo in 3D Ultrasound using PCA and an Ensemble of Heuristic, Atlas-based and Learning-based Classifiers Evaluated on the Rotterdam Periconceptional Cohort}

\titlerunning{Robust Alignment of the Human Embryo based on PCA}

\author{Nikolai Herrmann\inst{1,2}\equalcontrib (\texttt{n.herrmann@erasmusmc.nl}) \and Marcella C. Zijta  \inst{1,2,3}\equalcontrib  \and Stefan Klein\inst{2} \and R\'egine P.M. Steegers-Theunissen\inst{1} \and  Rene M.H. Wijnen\inst{4} \and Bernadette S. de Bakker\inst{3,4,5} \and Melek Rousian\inst{1}  \and Wietske A.P. Bastiaansen\inst{1,2}}

\institute{Department of Obstetrics and Gynecology, Erasmus MC, Rotterdam, The Netherlands \and Biomedical Imaging Group Rotterdam, Department of Radiology and Nuclear Medicine, Erasmus MC, Rotterdam, The Netherlands \and Department of Obstetrics and Gynecology, Amsterdam UMC, Amsterdam, The Netherlands \and Department of Pediatric Surgery, Erasmus MC, Rotterdam, The Netherlands \and Amsterdam Reproduction and Development Research Institute, Amsterdam, The Netherlands }

\def\codelink{\url{https://gitlab.com/radiology/prenatal-image-analysis/pca-3d-alignment}}

\authorrunning{N. Herrmann, M.C. Zijta et al.}

\maketitle

\begingroup
\renewcommand\thefootnote{+}
\footnotetext{These authors contributed equally to this work.}
\endgroup 
\vspace{-1.5em}

\begin{abstract}
Standardized alignment of the embryo in three-dimensional (3D) ultrasound images aids prenatal growth monitoring by facilitating standard plane detection, improving visualization of landmarks and accentuating differences between different scans. In this work, we propose an automated method for standardizing this alignment. Given a segmentation mask of the embryo, Principal Component Analysis (PCA) is applied to the mask extracting the embryo’s principal axes, from which four candidate orientations are derived. The candidate in standard orientation is selected using one of three strategies: a heuristic based on Pearson's correlation assessing shape, image matching to an atlas through normalized cross-correlation, and a Random Forest classifier. We tested our method on 2166 images longitudinally acquired 3D ultrasound scans from 1043 pregnancies from the Rotterdam Periconceptional Cohort, ranging from 7+0 to 12+6 weeks of gestational age. In 99.0\% of images, PCA correctly extracted the principal axes of the embryo. The correct candidate was selected by the Pearson Heuristic, Atlas-based and Random Forest in 97.4\%, 95.8\%, and 98.4\% of images, respectively. A Majority Vote of these selection methods resulted in an accuracy of 98.5\%. The high accuracy of this pipeline enables consistent embryonic alignment in the first trimester, enabling scalable analysis in both clinical and research settings. The code is publicly available at: \codelink.

\end{abstract}

\section{Introduction}
The first trimester of pregnancy is a crucial period for embryonic and fetal development, during which significant anatomical and physiological changes occur \cite{Carlson2013}. Throughout this time, the foundations for all major organ systems are established, making it a critical phase for monitoring embryonic and fetal growth and detecting congenital anomalies \cite{Bardi2022,Flierman2023}. Traditionally, pregnancies are monitored in utero using two-dimensional (2D) ultrasound by inspecting standard planes and obtaining several measurements from these planes \cite{ISUOG2023}. However, three-dimensional (3D) ultrasound imaging offers more detailed spatial information than traditional 2D images, enabling more comprehensive analysis of embryonic development \cite{jong-pleij2010}. For example, while the crown-rump length (CRL) is a standard 2D measurement in clinical practice, research has shown that the embryonic volume (EV) offers better insights into early development and early detection of congenital abnormalities \cite{Baken2017,baken2013first,martins2008first}.

To exploit the potential of 3D ultrasound, standardized alignment of the embryo is essential as the embryo can be located in any orientation and position within the womb. Without alignment of the embryo, this variability complicates consistent comparisons between images over time and across subjects. Clinically, standardized alignment of the embryo supports accurate identification of anatomical landmarks and standard planes for reproducible biometric measurements, enabling reliable assessment of growth and development \cite{bastiaansen_alignment,Yaqub2015}. A standard spatial orientation improves the visibility of subtle structural differences, supporting automated approaches to tasks like anomaly detection \cite{Qiu2017,Yaqub2015}. This consistency may be particularly beneficial for neural networks, which are typically not rotation invariant by design \cite{Goodfellow-et-al-2016}.

Most studies on alignment of the embryo and fetus have focused on the brain \cite{chen2012registration,kuklisova2013registration,moser2022bean,namburete2018fully}, while little work has addressed alignment of the entire embryo \cite{bastiaansen_alignment}. Alignment is typically achieved by either feature- or deep learning-based registration techniques, with many methods using auxiliary data such as segmentations \cite{kuklisova2013registration,namburete2018fully} or landmarks \cite{bastiaansen_alignment} to improve registration. Following this, we propose a two-step alignment approach: segmentation followed by alignment. Given the high degree of outer morphological symmetry around the mid-sagittal plane of embryos in neutral position \cite{ISUOG2023}, we explore rigid alignment. To achieve rigid alignment, Principal Component Analysis (PCA) offers a well-established approach for defining reference frames in 3D point clouds \cite{kasaei2020orthographicnet}. It has been successfully applied to align various human anatomical structures, including teeth \cite{nizu2024alignment} and the uterus \cite{bonevs2024automatic}.

Building on these insights, we propose an automated method that uses PCA to rigidly align the human embryo in 3D ultrasound images to a standardized orientation. Segmentations of the embryo are considered as 3D point clouds, in order to identify the principal axes of embryonic shape variation, resulting in multiple candidate orientations. To determine the correct orientation among them, we propose three selection strategies, either based on heuristics or machine learning. The performance of our alignment method is assessed using over 2000 3D ultrasound images acquired between 7+0 and 12+6 gestational age. 

\section{Method}
\subsection{Overview} 
\begin{figure}[H]
    \centering
    \includegraphics[width=0.9\linewidth]{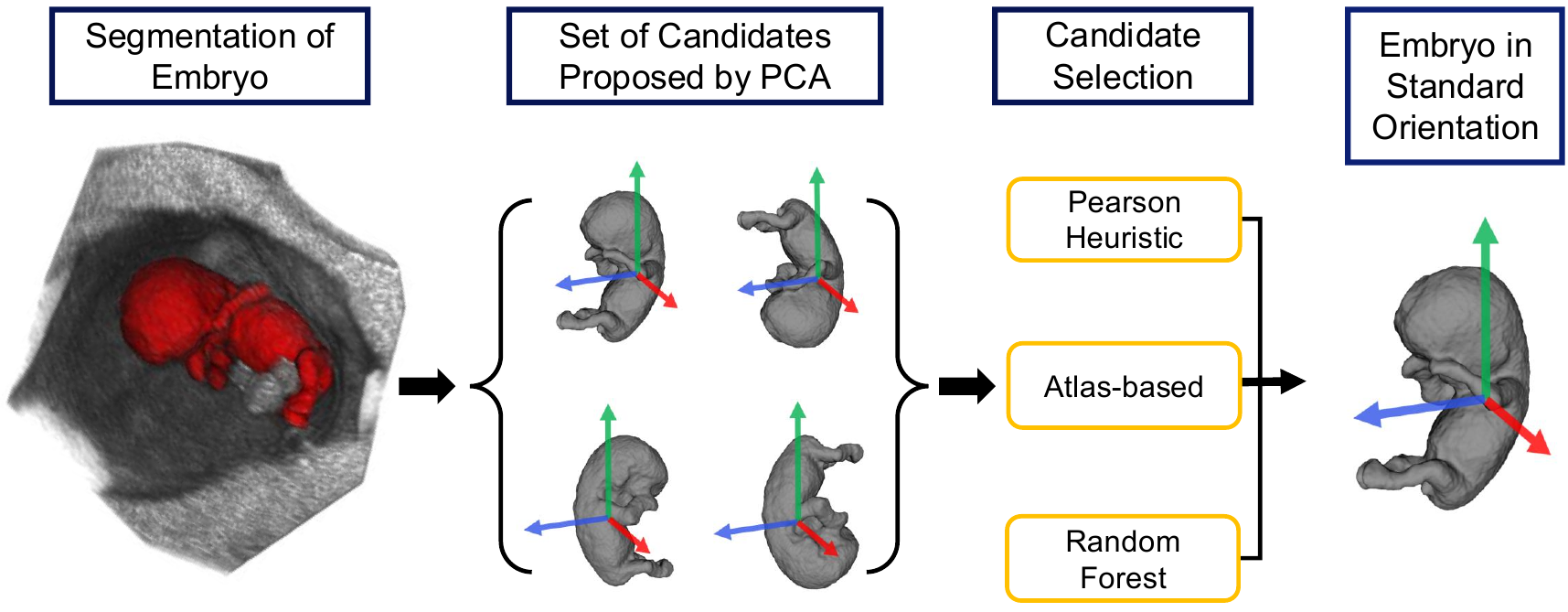}
    \caption{Overview of our alignment method with an example 3D ultrasound image acquired in the 10\textsuperscript{th} gestational week. Our method takes as input a 3D segmentation of the embryo and then calculates possible candidate images using Principal Component Analysis (PCA). Three selection methods assess all candidates and select the one in standard orientation. These are also combined in a Majority Vote.}
    \label{fig:method}
\end{figure}

Given a 3D ultrasound image $I$ with segmentation mask $S$ of the embryo, we propose to achieve spatial alignment by applying PCA to $S$ yielding a rotation matrix $U$ that rigidly transforms $I$ into standard orientation. Here, we assume a neutral embryonic position \cite{ISUOG2023}, as the 3D ultrasound volumes in our dataset were specifically acquired in a standardized manner to enable accurate CRL measurement in neutral pose. A caveat of PCA is the inherent sign ambiguity of its eigenvectors \cite{bro2008resolving,kasaei2020orthographicnet}, leading to four possible rotation matrices from which we derive four aligned candidate images. Following convention from previous work \cite{bastiaansen_alignment}, an embryo is considered to be in standard orientation when upright and facing to the left. We compare three different methods that examine all candidates and select the one in standard orientation. Figure \ref{fig:method} gives a schematic overview of our method. The code of our method is publicly available at \codelink.

\subsection{Principal Component Analysis}
The input for PCA is a 3D point cloud $X$ with its center of mass at the origin, constructed from the 3D coordinates of the non-zero voxels in $S$ \cite{kasaei2020orthographicnet}. PCA is applied to $X$ in order to capture the three most significant directions of variation in $S$. To calculate the principal components we used the implementation by \texttt{scikit-learn} \cite{scikit-learn} (Version 1.2.2) which uses single value decomposition (SVD) \cite{bro2008resolving}: $X = U \Sigma V^T$. Notably, the decomposition is not unique, as reflecting the singular vectors (columns of U and V) still yields a valid factorization, resulting in a set of $N=8$ possible rotation matrices $(U_i)_{i=1,\dots,N}$ due to independent reflection of the columns in U \cite{bro2008resolving}. Given that $U_i$ is an orthogonal matrix, and assuming a right-handed coordinate system, we can reduce to a set of $N=4$ distinct rotation matrices by ensuring that the sign of the determinant is positive: $\det(U_i) = 1$. Each rotation matrix $U_i$ can be applied to $I$ yielding a candidate image $C_i$. Candidate images were pre-processed by setting background voxels to zero and after rotation re-centered and padded such that the embryo's center of mass is in the center of the image. 

Package \texttt{scikit-learn} enforces a deterministic SVD output, based on the signs of the singular vectors, for code reproducibility. In preliminary experiments, we found that mirroring the first two principal components from the deterministic SVD output aligned the majority of images to standard orientation. We consider this the default candidate selection method. 

\subsection{Candidate Selection}
We investigated three different methods, that are independent of the PCA implementation, which select the candidate in standard orientation.

\begin{figure}[H]
    \centering
    \includegraphics[width=0.8\linewidth]{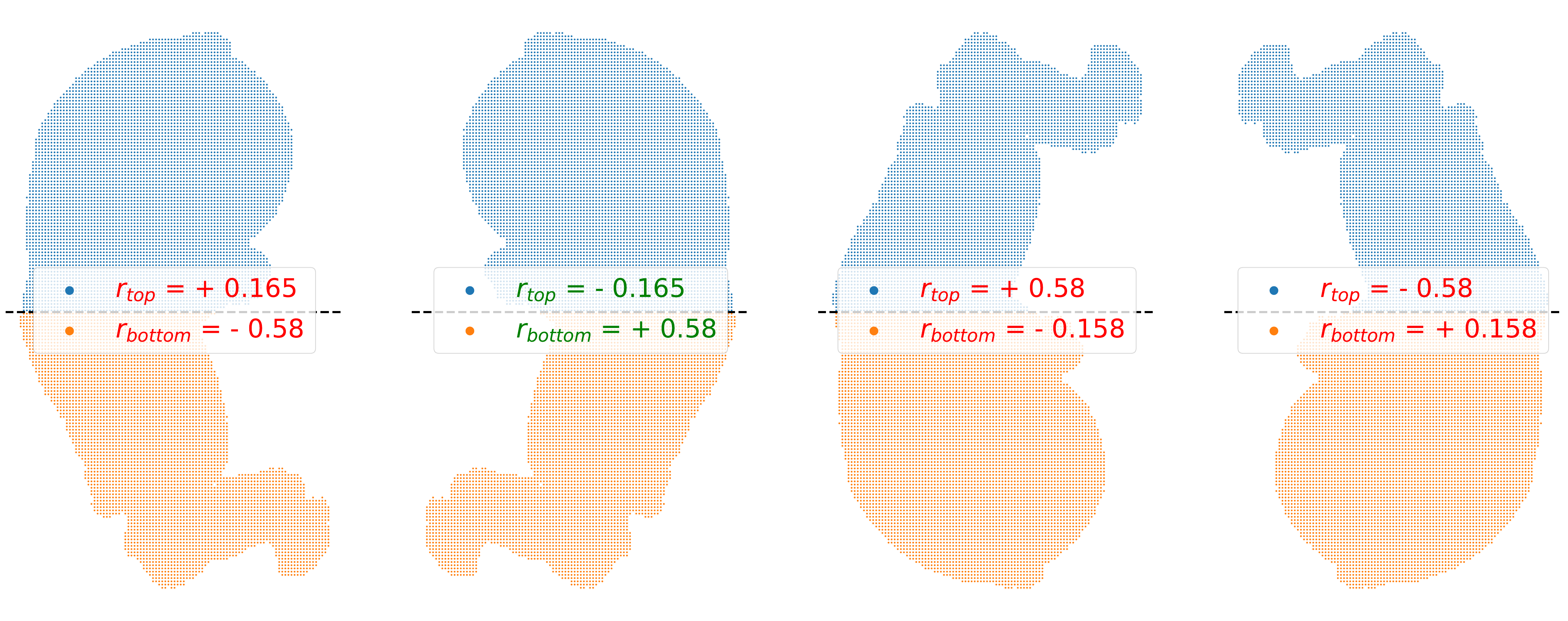}
    \caption{Example of a reduced 3D point cloud of a fetus from the 11\textsuperscript{th} gestational week. The Pearson correlation coefficients, $r_{top}$ and $r_{bottom}$, are computed for both the top and bottom section, respectively. The selection criteria for the Pearson Heuristic method, $|r_{bottom}| > |r_{top}|$ and $r_{bottom} > 0$, hold true in the second point cloud.}
    \label{fig:pearson}
\end{figure}

\textbf{Pearson Heuristic}
Following the works of Kasaei \cite{kasaei2020orthographicnet} and Nizu et al. \cite{nizu2024alignment} we examine the candidates' silhouette and calculate metrics that are indicative of orientation. The silhouette of candidate $C_i$ is obtained by transforming $X$ with $U_i$ and then projecting the point cloud onto a 2D plane, discarding the axis corresponding to the smallest eigenvalue (red arrow in Figure \ref{fig:method}). Resulting 2D point clouds, as shown in Figure \ref{fig:pearson}, are then split horizontally at the center in the direction of the largest eigenvalue, forming two sections. For both the top and bottom section the Pearson correlation coefficients, $r_{top}$ and $r_{bottom}$ respectively, are computed in order to assess shape \cite{kasaei2020orthographicnet}. Given a neutral embryonic position, an embryo has a curved body, hence, $r_{top}$ and $r_{bottom}$ should be opposing in sign. An embryo facing to the left will have a positive correlation in the bottom section and a negative correlation in the top section. The strength of the two correlations is assessed in order to ensure that the embryo is upright. Through preliminary experiments, we found that the section containing the embryos' legs gives a stronger correlation. Thus, the candidate in standard orientation is the one in which $|r_{bottom}| > |r_{top}|$ and $r_{bottom} > 0$ hold.

\textbf{Atlas-based}
To find which of the four candidates is most similar to an embryo in standard orientation we compute the normalized cross-correlation between each candidate and an atlas $A$ that has been manually put into standard orientation. Given the rapid morphological development within the first trimester we use a set of atlases acquired across different weeks. We adopt the set from previous work \cite{bastiaansen_alignment}: $(A_{i,j})_{i=1,\dots,8, j=8,\dots, 12}$ where $i$ represents the unique subject and $j$ the gestational week. We define the most appropriate atlas $A_{i^*,j^*}$ for the embryo present in image $I$, as the atlas most similar in EV \cite{Bastiaansen2025}: $i^*,j^* = \text{argmin}_{i, j} | \text{EV}(A_{i,j}) - \text{EV}(S)|$. Before calculating the cross-correlation score all four candidates were center-cropped and isotropically rescaled to a size of $64 \times 64 \times 64$, matching the dimensions of the available atlases \cite{bastiaansen_alignment}. The zoom factor for rescaling was set to $\sqrt[3]{\frac{M}{EV(S)}}$ where $M$ is the median EV of all atlases. Normalized cross-correlation was calculated using the \texttt{scipy} \cite{2020SciPy-NMeth} package (Version 1.14.1).

\textbf{Random Forest}
We trained a Random Forest classifier to identify standard orientation of candidate $C_i$ based on its mid-sagittal plane. We selected a Random Forest, as ensemble methods are generally more robust to noise \cite{breiman2001random}. The classifier was implemented using package \texttt{scikit-learn} \cite{scikit-learn} (Version 1.2.2). For pre-processing, candidate images were isotropically resized to $192 \times 192 \times 192$. The mid-sagittal plane was extracted at the central slice along the axis of the smallest eigenvalue, and the resulting 2D slice was flattened to obtain an input feature vector.

\textbf{Majority Vote}
Finally, to leverage the complementary strengths of the individual methods—Pearson Heuristic, Atlas-based, and Random Forest—we combined their predictions through majority voting. In cases where all methods disagreed, no candidate was selected and the image was marked as a failure.

\section{Data \& Experiments}
3D ultrasound images were included from the Rotterdam Periconceptional Cohort (Predict Study), conducted at the Erasmus MC, University Medical Center Rotterdam, The Netherlands \cite{Rousian_predict,Steegers_predict}. The aim of the cohort is to investigate the association between parental health and embryonic growth and development, during the periconceptional period. The 3D ultrasound images were acquired transvaginally, using Voluson E8 devices (GE Healthcare, Austria). Participants were at least 18 years old and had an ongoing singleton pregnancy of less than 10 weeks' gestational age.

For this study, we included data of 2848 3D ultrasound scans of 1368 pregnancies longitudinally acquired from 7+0 to 12+6 gestational age. Two raters (NH and MZ) independently performed quality control and excluded 172 images. Images were removed if both raters judged the original image to be of insufficient quality for biometric measurements — due to low resolution or noise — or if the entire embryo was not visible in the image. After quality control, we randomly split the dataset into 20\% for training and 80\% for testing, resulting in a training set of 510 images (261 pregnancies) and a test set of 2166 images (1043 pregnancies).

To apply PCA and generate the four orientation candidates, segmentations of the embryo were obtained using a previously trained nnU-Net model \cite{Bastiaansen2025}. The resulting PCA-candidates for both the train and test set were then independently reviewed by the raters, each selecting the candidate that matched standard orientation. If either rater identified that none of the candidates were in standard orientation, the image was marked as a PCA failure. 

Before training the Random Forest, PCA failures were removed from the training dataset (7 images, 2 pregnancies). Hyperparameter tuning was performed only on the training set using 5-fold cross-validation, the best model configuration used 200 trees and the Gini impurity splitting criterion while keeping all other \texttt{scikit-learn} hyperparameters default. After tuning, the entire training dataset was used to train the final Random Forest model.

Finally, to assess the performance of each selection method, accuracy was calculated on the test set. To statistically compare methods, we used the McNemar mid-$p$ test \cite{mcnemar1947note} to determine whether each method's accuracy differed significantly from the accuracy of the Majority Vote. The mid-$p$ variant was chosen as it is more sensitive than the standard McNemar test, especially when differences between methods are small and sample sizes are limited \cite{Fagerland2013}. Differences were considered statistically significant at $p<0.05$. In addition, we examined performance by gestational week to evaluate how each method performed across different stages of early development.

\section{Results}
\begin{figure}[H]
    \centering
    \includegraphics[width=0.9\linewidth]{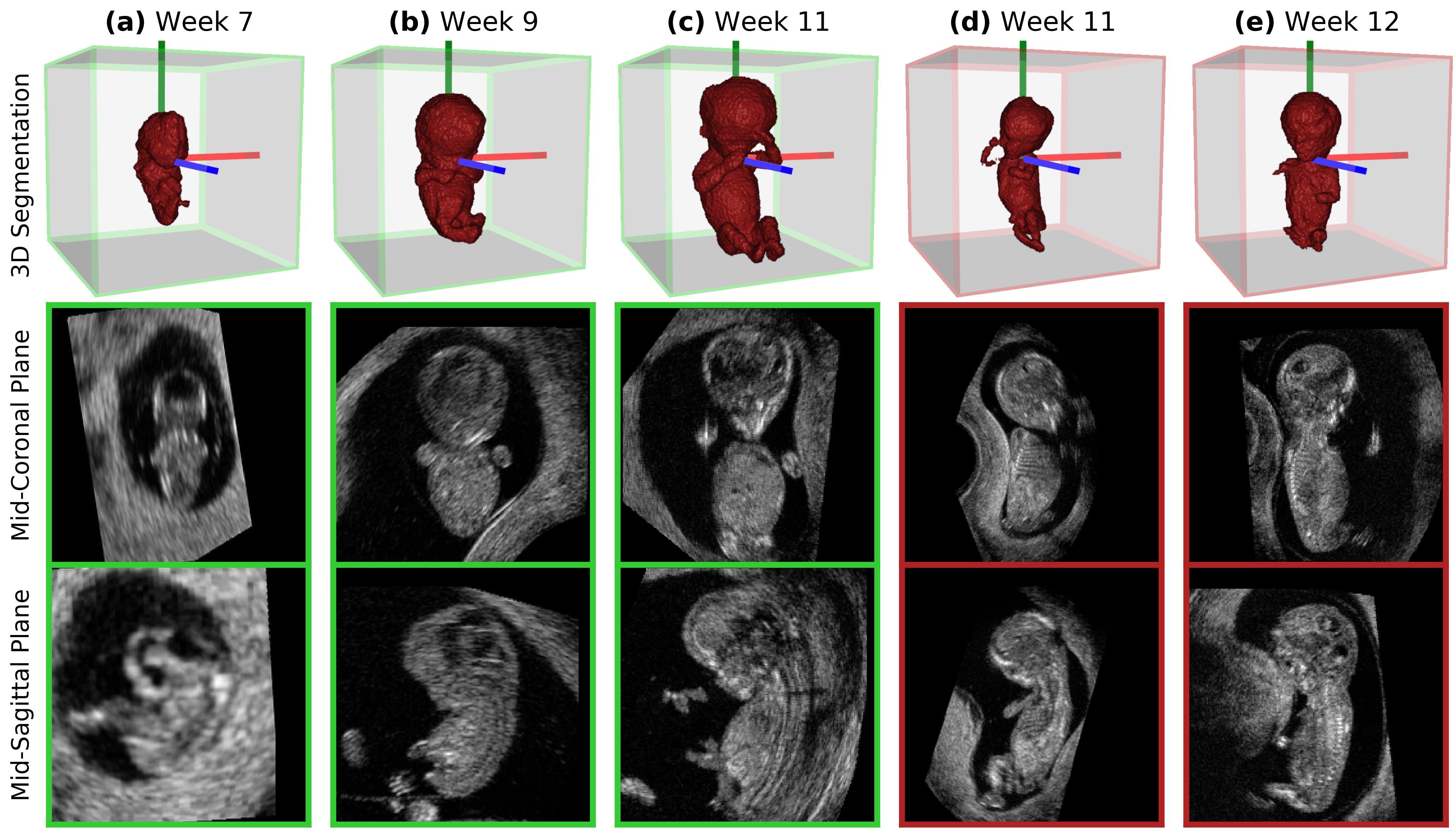}
    \caption{Columns \textbf{(a)} to \textbf{(c)} show correctly aligned test set images across gestational weeks, while columns \textbf{(d)} and \textbf{(e)} show failure cases to due non-neutral position and missing limbs in the segmentation, respectively. The mid-sagittal and mid-coronal planes as well as the 3D rendering are shown per column for each unique scan.}
    \label{fig:example_images}
\end{figure}

Of the 2166 images in the test set, PCA generated the correct standard orientation in 99.0\%. Figure~\ref{fig:example_images}\textit{a}-\textit{c} shows examples of embryos and fetuses of gestational week 7, 9, and 11 successfully aligned to standard orientation, with the three arrows indicating the direction of the principal components. The mid-coronal and mid-sagittal planes display key anatomical landmarks, such as the curvature of the embryonic body and the developing brain ventricles, confirming that the embryos are aligned to standard orientation \cite{ISUOG2023}. PCA failures were attributed to non-neutral embryonic positions, as illustrated in Figure~\ref{fig:example_images}\textit{d}, or errors in the segmentations, such as missing limbs (Figure~\ref{fig:example_images}\textit{e}). 

The accuracies of the different selection methods are summarized in Table \ref{tab:flip_results}. By taking the default PCA-candidate an accuracy of 86.3\% across all data was obtained. The individual selection methods all achieved accuracies greater than 95.0\%, with Majority Vote having the highest overall accuracy with 98.5\%. This was significantly better than all other methods except for Random Forest (98.4\%). The week-by-week analysis in Table~\ref{tab:flip_results} revealed performance differences across gestational weeks. Most PCA failures were in week 12 (93.2\%). In weeks 7 and 8, the default PCA candidate was in standard orientation in at most 43.6\% of the images, while in the later weeks its accuracy exceeded 93.2\%, comparable to the selection methods. The Pearson heuristic consistently achieved accuracies above 87.8\% but was significantly less accurate than Majority Vote in weeks 7 and 11. Atlas-based achieved the highest accuracy in week 7 (94.2\%) but significantly lower accuracies than Majority Vote in weeks 8 through 12. Random Forest and Majority Vote performed similarly with no statistical difference.

\vspace{-0.2em}
\vspace{-2.5em}

\begin{table}[H]
    \centering
    \begin{tabularx}{\textwidth}{
    >{\raggedright\arraybackslash}p{3.7cm}
    |>{\centering\arraybackslash}X
    |>{\centering\arraybackslash}X
    |>{\centering\arraybackslash}X
    |>{\centering\arraybackslash}X
    |>{\centering\arraybackslash}X
    |>{\centering\arraybackslash}X
    |>{\centering\arraybackslash}X
}
        \hline
        \textbf{Gestational week} & \textbf{All} & \textbf{7} & \textbf{8} &\textbf{9} & \textbf{10} & \textbf{11} & \textbf{12} \\
        Number of Test Images & 2166 & 172 & 241 & 667 & 265 & 600 & 221 \\
        \hline
        \multicolumn{7}{l}{\textit{Percentage of images with one candidate in standard orientation}} \\
        \hline
        PCA & 99.0 & 98.8 & 100.0 & 99.9 & 100.0 & 99.3 & 93.2 \\
        \hline
        \multicolumn{7}{l}{\textit{Accuracy of candidate selection method}} \\
        \hline
        PCA & 86.4* & 33.1* & 43.6* & 96.6*  & 99.6 & 99.2 & 93.2 \\
        PCA + Pearson Heuristic & 97.4* & 87.8* & 99.6 & 99.6 & 99.2 & 98.3* & 91.0 \\
        PCA + Atlas-based &  95.8* & 94.2 & 92.5* & 98.7* & 98.1* & 96.0* & 88.7* \\
        PCA + Random Forest  & 98.4 & 92.4 & 100.0 & 99.9 & 100.0 & 99.3 & 92.8 \\
        PCA + Majority Vote & 98.5 & 93.6 & 100.0 & 99.9 & 100.0 & 99.2 & 92.8   \\
        \hline
    \end{tabularx}
    \caption{Test set performance of Principal Component Analysis (PCA) and a comparison of selection methods for choosing the PCA candidate in standard orientation, across all and individual gestational weeks. * indicates a significant difference in accuracy compared to the Majority Vote ($p<0.05$).}
    \label{tab:flip_results}
\end{table}

\section{Discussion}
Standardized alignment of the embryo is essential for interpreting first-trimester 3D ultrasound, as embryonic positioning varies and affects both biometric assessment and further automated analyses. We developed an automated alignment method using PCA applied to segmentations, generating four candidates. In 99.0\% of the test images, the standard orientation was among these. To increase robustness, we included three selection methods based on distinct principles: The Pearson Heuristic leverages spatial geometry, the Atlas-based method uses image matching to a reference atlas, and the Random Forest relies on learned features. All methods outperformed the default PCA-candidate with Majority Vote achieving the highest accuracy (98.5\%). Although Random Forest and Majority Vote showed comparable performance across weeks, Majority Vote helps prevent systematic errors and ensures more reliable selection even if individual methods occasionally fail.

Slightly lower performance was observed in gestational weeks 7 and 12 compared to weeks 8–11. In week 12, alignment failures can be explained by segmentation errors as well as an increased occurrence of non-neutral fetal positions in this later stage of the first trimester \cite{Lchinger2008}. The segmentation errors likely stem from limited training data available for the segmentation algorithm for this particular gestational week \cite{Bastiaansen2025}. Since PCA is applied to the segmentation masks, its performance is highly sensitive to segmentation quality—for example, missing limbs can distort the computed principal components. In week 7, failures were typically caused by incorrect candidate selection, likely related to lower image resolution and quality due to the small embryo size. These observations suggest that segmentation completeness, neutral embryonic position, and overall image quality—beyond the current exclusion of low quality and noisy images—may be important considerations for future applications. 

A key strength of the proposed approach is that PCA requires no training and can be applied to segmentations of any image size, as it is not constrained by fixed input dimensions. Similarly, the Pearson Heuristic and Atlas-based methods require no training, with the latter relying only on a set of atlases. While the Random Forest classifier does require training, it is relatively lightweight compared to deep neural networks and was trained on a small dataset of $\sim 500$ images. PCA + Majority Vote takes less than 20 seconds on average per image on standard modern hardware, making it not only efficient but also practical for large-scale or time-sensitive applications. Overall, the method remains adaptable and easily transferable to other datasets and tasks.

While our alignment method demonstrates high accuracy, its performance was assessed only through visual inspection due to the absence of ground truth annotations. Consequently, minor deviations in orientation—on the order of a few degrees—may have gone unnoticed. Additionally, the method performs rigid alignment only, which cannot account for non-neutral embryonic positions. To address these challenges, future work could incorporate deformable alignment techniques that better capture natural anatomical variations and postural differences. Systematic accuracy evaluation could be introduced by using synthetic datasets with explicit orientations or expert-annotated images. Additionally, customizing heuristics based on gestational age or target anatomy may further improve alignment consistency across diverse datasets.

Standardized embryonic alignment in 3D ultrasound scans is essential to reliably compare anatomical features across embryos and their different developmental stages. For example, it can support clinical tasks such as standard plane detection; the mid-sagittal plane used for CRL measurement \cite{ISUOG2023} can be automatically extracted from the aligned embryo. Furthermore, it is a critical pre-processing step for population-based analyses such as atlas construction \cite{bastiaansen20254dhumanembryonicbrain,namburete2018fully}. Future work could build on our alignment method to develop tools for congenital anomaly detection, automated organ assessment, or cross-modality comparisons.

In conclusion, we developed an automatic alignment method for human embryos between gestational weeks 7 and 12 using 3D ultrasound imaging. Our approach achieved consistent standard orientation with an overall accuracy of 98.5\%. With such reliability, our full-body embryonic alignment method supports both clinical applications such as automatic biometric measurements and research efforts including longitudinal growth studies and population-level modeling. By providing a key pre-processing step for downstream analyses, this method contributes significantly to advancing early pregnancy assessment and paves the way for future innovations in prenatal imaging and diagnosis.

\begin{credits}
\subsubsection{\ackname} This publication is part of the project AI4AI with file number P22.017 of the research programme Perspectief which is (partly) financed by the Dutch Research Council (NWO) under the grant TTW-Perspectief 2022-2023. This preprint has not undergone peer review (when applicable) or any post-submission improvements or corrections. The Version of Record of this contribution is published in International Workshop on Preterm, Perinatal and Paediatric Image Analysis, and is available online at \url{https://doi.org/10.1007/978-3-032-05997-0_15}.
\end{credits}

\bibliographystyle{splncs04}
\bibliography{library_shorter}

\end{document}